\documentclass{article}

\usepackage{arxiv}

\usepackage[utf8]{inputenc} 
\usepackage[T1]{fontenc}    
\usepackage{hyperref}       
\usepackage{url}            
\usepackage{booktabs}       
\usepackage{amsfonts}       
\usepackage{nicefrac}       
\usepackage{microtype}      
\usepackage{lipsum}		
\usepackage{graphicx}
\usepackage{natbib}
\usepackage{doi}

\usepackage{amsthm}
\usepackage{floatrow}
\usepackage{multirow}
\usepackage[misc]{ifsym}
\usepackage{float}
\floatstyle{plaintop}
\restylefloat{table}
\usepackage{siunitx}
\usepackage{tabularx}
\usepackage{subcaption}
\usepackage[english]{babel}

\usepackage{amsmath}

\title{Forecasting Unobserved Node States with
spatio-temporal Graph Neural Networks}

\author{ \href{https://orcid.org/0000-0002-0515-7635}{\includegraphics[scale=0.06]{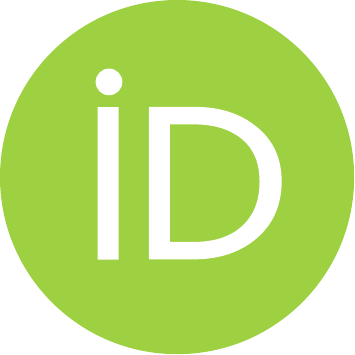}\hspace{1mm}Andreas Roth} \\
	Artificial Intelligence Group\\
	TU Dortmund\\
	Dortmund, Germany \\
	\texttt{andreas.roth@tu-dortmund.de} \\
	\And
	\href{https://orcid.org/0000-0002-9841-1101}{\includegraphics[scale=0.06]{orcid.pdf}\hspace{1mm}Thomas Liebig} \\
	Artificial Intelligence Group\\
	TU Dortmund\\
	Dortmund, Germany \\
	\texttt{thomas.liebig@tu-dortmund.de} \\
}

\renewcommand{\headeright}{}
\renewcommand{\undertitle}{}
\renewcommand{\shorttitle}{Forecasting Unobserved Node States}

\hypersetup{
pdftitle={Forecasting Unobserved Node States with
spatio-temporal Graph Neural Networks},
pdfsubject={machine learning, graph neural networks, spatio-temporal, imputation},
pdfauthor={Andreas Roth, Thomas Liebig},
pdfkeywords={machine learning, graph neural networks, spatio-temporal, imputation},
}

\begin{document}
\maketitle

\begin{abstract}
Forecasting future states of sensors is key to solving tasks like weather prediction, route planning, and many others when dealing with networks of sensors. 
But complete spatial coverage of sensors is generally unavailable and would practically be infeasible due to limitations in budget and other resources during deployment and maintenance.
Currently existing approaches using machine learning are limited to the spatial locations where data was observed, causing limitations to downstream tasks.
Inspired by the recent surge of Graph Neural Networks for spatio-temporal data processing, we investigate whether these can also forecast the state of locations with no sensors available.
For this purpose, we develop a framework, named Forecasting Unobserved Node States (FUNS), that allows forecasting the state at entirely unobserved locations based on spatio-temporal correlations and the graph inductive bias. FUNS serves as a blueprint for optimizing models only on observed data and demonstrates good generalization capabilities for predicting the state at entirely unobserved locations during the testing stage. 
Our framework can be combined with any spatio-temporal Graph Neural Network, that exploits spatio-temporal correlations with surrounding observed locations by using the network's graph structure. Our employed model builds on a previous model by also allowing us to exploit prior knowledge about locations of interest, e.g. the road type.
Our empirical evaluation of both simulated and real-world datasets demonstrates that Graph Neural Networks are well-suited for this task.
\end{abstract}


\section{Introduction}
\begin{figure}[tb]

  \centering
  \includegraphics[width=0.45\textwidth]{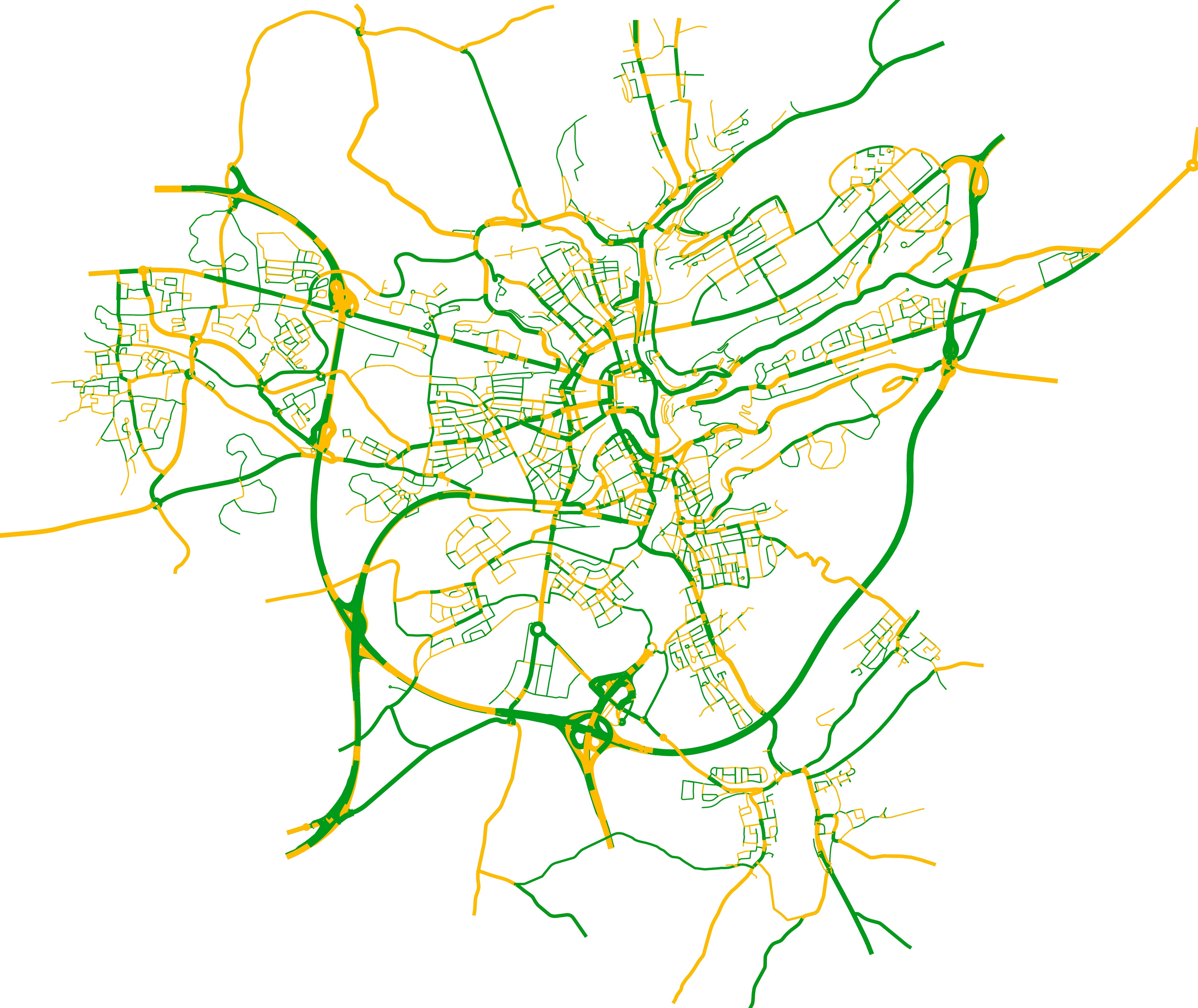}
  \caption{The road network of Luxembourg. We consider the scenario that sensors on green colored roads record time series data, and the traffic should be forecast for yellow colored roads, at which no sensors are available. In this illustration, sensors are placed on 50\% of all roads.}
  \label{fig:motivation}
\end{figure}
Consider the case of planning a route or estimating arrival times for your vehicle. To precisely calculate proper solutions, situational-aware algorithms require an estimation of the traffic state across the entire road network~\cite{LIEBIG2017258}. In crowded cities with rapid changes in congestion, algorithms benefit from knowing whether the future state of traffic will change during driving~\cite{derrow2021eta}.
Placing sensors across at all spatial points of interest is not feasible in practice since the corresponding costs for hardware, power and maintenance grow quickly.
While data is often available for main roads, e.g. highways, data for side road segments is sparsely available at best. Even for main roads, data is only available at selected positions.
Despite not having collected data at these points, we are interested in the state at these unobserved locations to support downstream tasks effectively. Fig.~\ref{fig:motivation} illustrates the case when $50\%$ of the roads are observed, but the remaining roads are not.
Similar dynamics happen not only in the case of traffic but for other domains where spatio-temporal data is collected, e.g. ground stations in weather forecasting.

Estimating the state at these unobserved locations from learning spatial dependencies from available data at surrounding locations seems promising.
But despite this being practically relevant, to the best of our knowledge, this problem has not yet been considered in the literature of machine learning approaches.
We find that the reason for the lack of consideration of this task results from most methods not being applicable for this task.
Traditional methods are typically usable for either spatial predictions, like Gaussian Processes~\cite{banerjee2008gaussian} or k-Nearest Neighbors (kNN)~\cite{batista2002study}, or for time series forecasting, e.g. autoregressive methods~\cite{dickey1979distribution}. At a future timestep neither simultaneous spatial observations nor any previous temporal observations at these locations are available.
Even most deep learning approaches cannot be applied for this task. A Multilayer perceptron (MLP)~\cite{rumelhart1985learning} considers neither spatial nor temporal dependencies and a Recurrent Neural Network (RNN)~\cite{10.1162/neco.1997.9.8.1735} makes only use of spatial data. While Convolutional Neural Networks (CNNs)~\cite{lecun1989backpropagation} are designed for spatial processing, they have been combined successfully with RNNs to process both spatial and temporal data~\cite{qiu2017learning,chao2018rethinking,xu2015learning}. But these are still not applicable for this task because employed convolution requires data that is regular, which is typically not the case for sensor networks. 

Recently, Graph Neural Networks (GNNs)~\cite{kipf2016semi} have seen a rise in popularity due to them generalizing the convolutional operation to non-euclidean data structures.
GNNs process irregular spatially connected and can, similarly to CNNs, be combined with temporal methods to process irregular spatio-temporal data~\cite{jiang2021graph,yu2017spatio,wu2020connecting,Chen_Li_Teo_Zou_Wang_Wang_Zeng_2019}. 
Expanding potential domains of application for GNNs has been a recent topic of interest~\cite{cini2021filling,bessadok2021graph,liao2021review}.
GNNs achieved remarkable progress for the task of predict future traffic states~\cite{jiang2021graph,Chen_Li_Teo_Zou_Wang_Wang_Zeng_2019}. However, current methods are designed to predict the future state at available sensor's respective geospatial locations. 
Our goal is to forecast these future states at locations where no sensors are placed. We expect a strong correlation between nearby sensors~\cite{tobler1970computer}.
Take crossroads in the traffic scenario as an example: Observing cars on three roads should allow a deduction of the number of cars on the fourth road.
Even though adjacent roads influence each other, the difference may be largely based on locational properties, e.g. highway and off-ramp.
The main goal of this paper is to show that graph neural networks are the first family of algorithms that is effective for this task.

Towards this goal, we propose a framework for Forecasting Unobserved Node States (FUNS) which optimizes any spatio-temporal GNN on the set of observed nodes and allows generalization to entirely unobserved nodes during testing and inference. 
This gets enabled by using the graph inductive bias as prior knowledge for connecting spatially related locations.
Completing the framework, we propose the FUNS-Network (FUNS-N), an adaptation of a previous spatio-temporal GNN for the imputation of temporal sensor failures~\cite{cini2021filling} that also incorporates additional prior knowledge about spatial locations, e.g., the road type.
model that exploits spatio-temporal correlations between nearby locations to predict.  

We summarize our key contributions as follows:
\begin{itemize}
    \item We introduce the unexplored task of forecasting \textbf{future} states of unobserved nodes and detail the connections to the previous tasks of forecasting and imputation.
    \item We introduce FUNS, a framework that allows the optimization and generalization of any spatio-temporal Graph Neural Networks for this task.
    \item Our empirical evaluation demonstrates that Graph Neural Networks are the first effective approach for this task.
\end{itemize}

We start by formally formulating our problem task in Sec.~\ref{sec:problem} and outlining similarities and differences to related approaches in Sec.~\ref{sec:related}. We then introduce our FUNS framework and detail how any model can be optimized in conjunction to generalize the forecast unobserved node states, in Sec.~\ref{sec:forecasting}. Following this, we construct our model for this task in Sec.~\ref{sec:fun-n}.
We evaluate our framework on a large-scale synthetic dataset simulating traffic for the entire state of Luxembourg and on METR-LA, which contains real-world highway data from Los Angeles, in Sec.~\ref{sec:experiments}. We conclude our work with potential prospects for future work in Sec.~\ref{sec:conclusion}.
\section{Problem Formulation}
\label{sec:problem}
In this section, we formally state our problem task and detail arising challenges.
We model observed and unobserved spatial locations jointly as nodes $\mathcal{V} = (v_1,\dots,v_n)$, with $|\mathcal{V}| = n$ being the number of nodes, of a graph $G = (\mathcal{V},\mathcal{E})$. The set of edges $\mathcal{E}$ consists of all node pairs $(v_i,v_j)$ that are spatially connected. The neighborhood of a node $v_i$ is defined as the set $N_i = \{j | (v_j,v_i) \in \mathcal{E}\}$ of node indices having an edge pointing towards $v_i$.

Contrary to related tasks, our framework assumes that sensors are only available for a subset of nodes $\hat{\mathcal{V}} \subset \mathcal{V}$ with $|\hat{\mathcal{V}}| = \hat{n}$. These sensors record $d$ measurements at $T$ discrete time steps, leading to a feature matrix $\hat{\mathbf{X}} \in \mathbb{R}^{T \times \hat{n} \times d}$. 
The set of unobserved nodes is labeled as the complement $\hat{\mathcal{V}}^C = \mathcal{V} \setminus \hat{\mathcal{V}}$ and the unknown true states at these locations with $\hat{\mathbf{X}}^C \in \mathbb{R}^{T \times (n-\hat{n}) \times d}$. 
We index a matrix with $\mathbf{X}^{[P:Q]} \in \mathbb{R}^{(Q-P) \times n \times d}$ to denote all time steps from $P$ to $Q$ and $\mathbf{X}^{[R]} \in \mathbb{R}^{1 \times n \times d}$ to denote the state at an individual time step.
Here, we assume the sensor measurements to be available uninterrupted for all $T$ time steps. Our goal is to find a function
\begin{equation}
\label{eq:target}
    f_{\theta}(\hat{\mathbf{X}}^{[0:Q]}, \mathcal{E}) = \hat{\mathbf{X}}^{C[Q+t]}
\end{equation}
that predicts the unobserved states for a future time step $Q+t$ by adjusting the learnable parameters $\theta$ of $f_{\theta}$. 
While any algorithm cannot use sensor measurements from nodes $\hat{\mathcal{V}}^C$, we assume the full graph structure $\mathcal{E}$ for observed and unobserved nodes to be known. To the best of our knowledge, this task has not been considered before.

We additionally allow sensor-independent information about the graph, which we call static node labels $\mathbf{L} \in \mathbb{R}^{n \times k}$, e.g., the road type for traffic data. The special challenge rises from the facts that at the considered time step $Q+t$ spatial data and at the considered nodes $\hat{\mathcal{V}}^C$ historical data is not available. Therefore, the combined spatial and temporal dynamics need to be learned and generalized.

Next, we clarify the differences between our problem task and related problem formulations in the literature.
\section{Related Work}
\label{sec:related}
While forecasting unobserved nodes has not been considered before, it is closely related to two widely known tasks. Our task is a mix of temporal forecasting, that potentially also considers spatial correlations, and the imputation of missing values. We will describe both tasks and connected approaches next.   

\subsection{Spatio-temporal forecasting}
Forecasting the future state of a graph based on fully available past states is closely related to our problem. The major difference is that the feature matrix $\mathbf{X} \in \mathbb{R}^{T \times n \times d}$ contains uninterrupted observations for all considered nodes. Therefore the goal is to find a function
\begin{equation}
    g_{\omega}(\mathbf{X}^{[0:Q]}, \mathcal{E}) = \mathbf{X}^{[Q+t]}
\end{equation}
with adjustable parameters $\omega$ that forecasts a future state $\mathbf{X}^{[Q+t]} \in \mathbb{R}^{n \times d}$ based on their respective past observations and spatial relationships encoded by their set of edges $E$. 
Classical approaches for statistical learning such as autoregressive integrated moving average (ARIMA)~\cite{williams2003modeling} already perform poorly in this task due to them not capturing spatial relationships. 
Recently, approaches based on spatio-temporal GNNs show state-of-the-art performance on various benchmark datasets~\cite{https://doi.org/10.1111/tgis.12644}. 
Guo et al.~\cite{guo2019attention} found that approaches based on deep learning outperform statistical methods especially for long-term predictions. 
All methods combine spatial graph convolutions with temporal operations known from sequence processing, like recurrent~\cite{Chen_Li_Teo_Zou_Wang_Wang_Zeng_2019}, convolutional~\cite{yu2017spatio}, or attentional~\cite{guo2019attention} modules.
While the main benchmark datasets contain traffic data, spatio-temporal GNNs were also proposed for other domains, including the prediction of solar power, electricity consumption prediction and daily exchange rates~\cite{wu2020connecting}, and weather forecasting~\cite{ma2022histgnn}. As our approach is agnostic to the domain, our framework can be applied here as well.

As all of these methods have not been applied to include the forecast of unobserved node states, it is unclear how these would perform. Our work is designed to examine the potential of adapted spatio-temporal GNNs for that task.

\subsection{Imputation}

Imputation in general is the task of dealing with missing values and incomplete data~\cite{schafer1997analysis}. 
In the context of spatio-temporal imputation, this translates to some of the observations being missing at some of the nodes, potentially for consecutive time steps~\cite{little2019statistical}. Missing observations are typically only short-term, with the motivation being dysfunctional sensors or network connection losses.
The task is to restore all of those missing values in available data.
Formally, we have a feature matrix with missing observations $\Tilde{\mathbf{X}} \in \mathbb{R}^{T \times n \times d}$ available and want to find a mapping
\begin{equation}
    h_{\psi}(\Tilde{\mathbf{X}}, \mathcal{E}) = \mathbf{X}\, ,
\end{equation}
that reconstructs our true feature matrix $\mathbf{X}$ by also considering the spatial structure $\mathcal{E}$.
The function $h_{\psi}$ has learnable parameters $\psi$.
Several approaches have been presented on the side of traditional machine learning, e.g. Gong et al.~\cite{1041310} impute missing values based on observations at the $k$ nearest neighbors.
Liebig et al.~\cite{liebig2012pedestrian} make use of a Gaussian Process Regression to predict the state at unobserved locations. Matrix completion algorithms (e.g.,~\cite{gross2011recovering,candes2010power}) cannot be used here because nodes are entirely unobserved.

Deep learning for spatio-temporal data was found to be effective for this task~\cite{yoon2018gain,liu2019naomi}, though these do not consider graph-structured data.
GNNs were explored for spatial imputation without a temporal aspect in the data~\cite{cini2021filling,you2020handling,spinelli2020missing}. These approaches deal with spatial imputation of nodes that have partially missing data. Closely related is the task of spatial interpolation for which Appleby et al.~\cite{appleby2020kriging} propose Kriging Convolutional Networks that are based on GNNs for interpolation to unknown locations. Similarly, Wu et al.~\cite{wu2020inductive} are applied to spatial interpolation by sampling random subgraphs, that can be used to introduce nodes at arbitrary locations.
One approach that considers the imputation of temporarily missing values in spatio-temporal data was proposed by Cini et al.~\cite{cini2021filling}. Here, a bidirectional model combines graph convolution and recurrent units to complete a matrix with an underlying graph structure.

None of these approaches is directly applicable to the task of predicting future unobserved node states.
The majority of these approaches does not even consider the prediction of present states of entirely unobserved nodes by considering spatio-temporal relations.

\section{FUNS Framework}
\label{sec:forecasting}

In our approach, we combine the spatial and temporal dependencies in the data to predict future states of unobserved nodes. We also use the same parameters for each node, so we can generalize to entirely unobserved nodes during testing. We start by detailing our Forecasting Unobserved Node States (FUNS) framework, which can be used to optimize any appropriate model. We then proceed to present our FUNS-Network that is adapted specifically for this task.

\subsection{Optimization}
The goal of FUNS is to optimize parameters $\theta$ of the function $f_{\theta}$ presented in~\eqref{eq:target}.
Our optimization procedure needs to allow generalization from the subset of available observations at nodes $\hat{\mathcal{V}}$ to all remaining nodes $\hat{\mathcal{V}}^C$.
FUNS accomplishes this by treating some of the available nodes as unobserved during training and therefore simulating the process desired in testing.
We mask part of the observed nodes so they are indistinguishable to unobserved nodes as an input to the model.
Models used in combination with FUNS should make use of shared parameters for all nodes, so that the model also processes these nodes in the same way.
Then during testing, we use all available observations to predict values for unobserved nodes. 
We split our available nodes $\hat{\mathcal{V}}$ accordingly into an input set $\hat{\mathcal{V}}_{\mathrm{in}}$ and an optimization set $\hat{\mathcal{V}}_{\mathrm{opt}}$ with $\hat{\mathcal{V}}_{\mathrm{in}} \cup \hat{\mathcal{V}}_{\mathrm{opt}} = \hat{\mathcal{V}}$ and $\hat{\mathcal{V}}_{\mathrm{in}} \cap \hat{\mathcal{V}}_{\mathrm{opt}} = \emptyset$.

For masking some of the nodes, we introduce the notation of a mask vector for a subset $\hat{\mathcal{V}}$ of nodes
\begin{equation}
\mathbf{m}^{({\hat{\mathcal{V}}})}_i=
\begin{cases}
  1, & \text{if}\ v_i \in \hat{\mathcal{V}} \\
  0, & \text{otherwise}
\end{cases}
\end{equation}
that contains binary values for the presence of each node $v_i \in \mathcal{V}$ in the subset $\hat{\mathcal{V}}$. Multiplying the feature matrix with a mask, that is broadcasted along the time and feature dimension, allows us to only consider the designated input nodes for our model.
As usual for sequential data, we additionally split our data  $\hat{\mathbf{X}} \in \mathbb{R}^{T \times \hat{n} \times d}$ along the temporal dimension into a training sequence $\hat{\mathbf{X}}^{[0:P]} \in \mathbb{R}^{P \times \hat{n} \times d}$, a validation sequence $\hat{\mathbf{X}}^{[P+1:Q]} \in \mathbb{R}^{Q-P \times \hat{n} \times d}$ and a test sequence $\hat{\mathbf{X}}^{[Q+1:T]} \in \mathbb{R}^{T-Q \times \hat{n} \times d}$ with $P < Q < T$ and $T$ being the number of total available time steps.

During training, we randomly sample subsequences $\hat{\mathbf{X}}^{[V:W]}$ of the training sequence with $V < W < P$ and allow any task-specific loss function $l$ to calculate the loss 
\begin{equation}
    \mathcal{L} = l(f(\hat{\mathbf{X}}^{[V:W]} \odot \mathbf{m}^{(\hat{\mathcal{V}}_{\mathrm{in}})},\mathcal{E}) ,\hat{\mathbf{X}}^{[V+t:W+t]})\, ,
\end{equation}
where $\odot$ denotes the Hadamard product. 
We improve generalization of our framework by two things: We predict the state for all time steps, even when no temporal context is available for the initial steps. And we calculate the loss using not only the masked nodes, but all available nodes. We find this optimization procedure to be connected to semi-supervised node classification~\cite{kipf2016semi}. Generalization of that method was found to be successful, even for new graphs~\cite{hamilton2017inductive}, so we expect similar generalization capacities here. Parameters $\theta$ are optimized using gradient descent with Backpropagation through time (BPTT)~\cite{werbos1988generalization,robinson1987utility}.

\subsection{Testing}
When applying our framework in real-world applications or evaluating its performance for comparison with other approaches, a couple parts differ from the optimization stage.
We use all available observations as input to our function $f_\theta$ without any masking. We also do not use fixed-length subsequences, but the entire prior sequence. This allows us to make predictions based on the entire available temporal context.
The test loss
\begin{equation}
    \mathcal{L}_{\mathrm{test}} = l(f(\hat{\mathbf{X}}^{[Q+1:T-t]},\mathcal{E}) \odot \mathbf{m}^{(\hat{\mathcal{V}}^C)},\mathbf{X}^{[Q+1+t:T]} \odot \mathbf{m}^{(\hat{\mathcal{V}}^C)})\, .
\end{equation}
is calculated only using the test sequence and test nodes $\hat{\mathcal{V}}^C$. The same process is also applied for the validation data.
Our function $f_{\theta}$ should therefore exploit the spatio-temporal correlations in the data while sharing processing and parameters across different nodes. In the following, we describe our choice for $f_{\theta}$, the FUNS-Network.
\section{FUNS-Network}
\label{sec:fun-n}
Since we are interested in showing that GNNs in general are well-suited for this task, we only conduct slight adaptations from previously successful models.
Our FUNS-Networks (FUNS-N) combines the tasks of spatio-temporal forecasting and spatio-temporal imputation, therefore any successful models from these domains could serve as a starting point. 
We chose to modify the Graph Imputation Network (GRIN)~\cite{cini2021filling}, a method that was successfully applied for the imputation task of temporarily sensor failures. They alternate Graph Recurrent Units (GRU)~\cite{cho2014learning} and graph convolutions using past and present observations to impute missing values in historic data.
We start by introducing GNNs and explain how FUNS-N uses them for spatial processing.

\subsection{Spatial Processing}
Given the large correlation between spatially connected nodes, we are interested in a method that deals with graph structures and learns the dependencies from data.
While convolutions are a very successful operation for combining spatial information in grid-structured data like images or sequences~\cite{he2016deep}, processing non-euclidean data like graphs is more challenging. A convolution operation on a graph needs to be able to deal with different numbers of neighbors and no ordering of neighbors being available. 

We make use of GNNs, specifically message passing neural networks (MPNNs)~\cite{gilmer2017neural}, that are currently a popular paradigm to achieve a convolution-like operation on graphs~\cite{gilmer2017neural}. Even though most GNNs follow the same structure, the details of the operations are still evolving~\cite{chen2020simple,brody2022how,roth2022transforming}.
The general framework utilizes the local neighborhood $N_i$ of each node to find updated node states
\begin{equation}
    \mathrm{MPNN}(\mathbf{H},\mathcal{E})_i = \psi(\mathbf{h}_i, \bigoplus_{j \in N_i} \omega(\mathbf{h}_i,\mathbf{h}_j) )\, ,
\end{equation}
where $\mathbf{H} \in \mathbb{R}^{n \times d}$ denotes a feature matrix of $n$ nodes and $d$ features and $\mathbf{h}_k \in \mathbb{R}^{d}$ denotes the feature vector of node $k$.
Here, $\omega$ denotes a message function that combines a neighboring state $\mathbf{h}_j$ with the root state $\mathbf{h}_i$. These messages are aggregated for all neighbors of a node by some permutation invariant aggregation operator $\bigoplus$. As a final step to update the node state, the aggregated messages are combined with the prior state of the node with another function $\psi$.
Note, that time is not considered in MPNNs and this operation is executed independently for each time step.

In our case, neighboring locations typically have different impacts on the future state of the root node, e.g. staying on a highway or leaving for an exit. Similarly, unavailable information from unobserved nodes should be mostly suppressed for message-passing operations.
Our idea is, that this prior knowledge should get used to determine which and how much information gets passed along an edge.
Therefore we adaptively weight the connection strength between pairs of nodes $(v_i,v_j)$ depending on static node labels, mask values, and the learned hidden state.

Many recently presented MPNNs, e.g. GCN~\cite{kipf2016semi}, GraphSAGE~\cite{hamilton2017inductive}, are unable to treat neighbors differently. However, MPNNs using the attention mechanism satisfy this property, e.g. Graph Attention Networks (GAT)~\cite{velickovic2017graph} and their more expressive version GATv2~\cite{brody2022how}, which we use for all MPNNs in this work. 
These make use of the attention mechanism~\cite{vaswani2017attention} to calculate pairwise coefficients $c_ij$ that weight all neighboring features of a node $v_i$. This allows us to control the flow by integrating external information we have about the underlying graph structure independently of the placement of our sensors.

\subsection{Architecture}
Contrary to GRIN, our framework does not allow future states to influence the prediction of past steps. We also show, how static node labels are considered to improve predictions.

We iterate through all time steps $t \in \{0,\dots,T\}$ and generate forecasts at each time step based on observed data from all previous time steps, so our approach can be used directly for real-time use cases.
We start by mapping our static node features, namely the binary mask $\mathbf{m} \in \{0,1\}^{n}$, its binary complement $\mathbf{\bar{m}} \in \{0,1\}^{n}$, and static node labels $\mathbf{L} \in \mathbb{R}^{n \times z}$ to a static state
\begin{equation}
    \mathbf{S} = \psi(\mathbf{m}||\mathbf{\bar{m}}||\mathbf{L})
\end{equation}
using the concatenation operation $||$ and a multilayer perceptron (MLP) $\psi$. We denote all available observations with $\mathbf{X}^{t} \in \mathbf{R}^{n \times d}$, where the features of nodes with no observations are set to zero. We then use all available information to approximate the current state at missing locations 
\begin{equation}
    \mathbf{Z}^t = MPNN(\mathbf{X}^t||\mathbf{H}^{t-1}||\mathbf{S})
\end{equation}
using the static state $\mathbf{S}$, $\mathbf{X}^{t}$, and our previous hidden state $\mathbf{H}^{t-1} \in \mathbb{R}^{n \times h}$, which is initialized as constant. We only use this approximation at unobserved locations, so we fill in the gaps of the feature matrix using the mask 
\begin{equation}
    \hat{\mathbf{X}}^t = \mathbf{m} \odot \mathbf{X}^t + \mathbf{\bar{m}} \odot \mathbf{Z}^t\, .
\end{equation}
At each time step, the input features are the concatenation $\mathbf{F}^{t} = \hat{\mathbf{X}}^t||\mathbf{S}$. We start by processing our input features with a temporal operation. In our case, this is a GRU 
\begin{equation}
    \mathbf{H}^t = \mathrm{GRU}(\mathbf{F}^t,\mathbf{H}^{t-1})
\end{equation}
that uses MPNNs as gates, as has been used before~\cite{seo2018structured,cini2021filling}. In addition to the input features $\mathbf{F}^t \in \mathbb{R}^{n \times d}$ at the current time step $t$, the internal hidden state $\mathbf{H}^{t-1}$ storing temporal dependencies is used as additional input. The operations inside the GRU are as described in~\cite{seo2018structured}:
\begin{equation}
    \mathbf{R}^{t} = \sigma(\mathrm{MPNN}(\mathbf{F}^t||\mathbf{H}^{t-1},E))
\end{equation}
\begin{equation}
    \mathbf{U}^{t} = \sigma(\mathrm{MPNN}(\mathbf{F}^t||\mathbf{H}^{t-1},E))
\end{equation}
\begin{equation}
    \mathbf{C}^{t} = \mathrm{tanh}(\mathrm{MPNN}(\mathbf{F}^t||\mathbf{R}^{t}\odot \mathbf{H}^{t-1},E))
\end{equation}
\begin{equation}
    \mathbf{H}^{t} = \mathbf{U}^{t} \odot \mathbf{H}^{t-1} + (1-\mathbf{U}^{t}) \odot \mathbf{C}^{t} 
\end{equation}
where $\sigma$ denoting the sigmoid activation function.
Initially, two gates are computed, namely the reset gate $\mathbf{R}^{t}$
determining which information from the hidden state to drop, and the update gate $\mathbf{U}^{t}$ is used for controlling which parts of the state to update. The candidate state $\mathbf{C}^{t}$ contains information that is potentially getting kept. Finally, the new hidden state $\mathbf{H}^{t}$ is calculated as a combination of the candidate state and the previous hidden state.

Even though this temporal step considers, spatial relations by using MPNNs as gates, we use another MPNN
\begin{equation}
    \mathbf{P}^{t} = \mathrm{MPNN(\mathbf{H}^{t}||\mathbf{F}^{t},E)}
\end{equation}
before making our prediction
\begin{equation}
    \mathbf{\hat{Y}}^{t} = \phi(\mathbf{P}^{t})\, .
\end{equation}
for our desired time step using an MLP $\phi$.

We use these predictions $\mathbf{\hat{Y}}^{t}$ for all time steps $t$ to calculate the loss as introduced in Section~\ref{sec:forecasting} to optimize all parameters using gradient descent.   
\section{Experiments}
\label{sec:experiments}
We evaluate the FUNS framework and our FUNS-Network jointly on both a real-world dataset and a simulated dataset, which has prior knowledge about locations available.
Our approach is designed to benefit from a detailed graph structure and additional static information about nodes that do not depend on sensors. We also need some available evaluation data for comparison.
One dataset, that has these information available comes from the Simulation of Urban MObility (SUMO)~\cite{SUMO2018}, which simulates microscopic traffic for road networks. 
We are using the Luxembourg SUMO Traffic (LuST) Scenario~\cite{codeca2017luxembourg}, modeling traffic for all roads of the state of Luxembourg. As these fine-grained details are not available for real-world datasets, we also  evaluate our approach for the case of little available information with METR-LA~\cite{li2018dcrnn_traffic}.

\subsection{LuST Scenario}
LuST consists of all roads in Luxembourg and their connections, with detailed geometric and road-specific properties, like the type of road or the number of lanes. Vehicles are simulated individually with realistic driving behavior across the road network. As the set of nodes, we use all $5779$ available roads, for which we observe all passing cars. We use the density and average speed aggregated over five-minute intervals. 
This leads to more than $800\ 000$ available observations spread over $294$ aggregated time steps, resulting in the feature matrix $\mathbf{X} \in \mathbb{R}^{294 \times 5779 \times 2}$.
The set of edges $\mathcal{E}$ is constructed as all pairs of roads that are directly connected. 
As static node labels, we use the road-specific properties road length, maximum allowed speed, and road type, which are each encoded as one-hot vectors. This results in the label matrix $\mathbf{L} \in \mathbb{R}^{5779 \times 13}$.

\subsubsection{Live prediction}
\begin{table*}[htbp]
\caption{MSE for the prediction of unobserved node states with $t=0$ for SUMO}
\begin{center}
\begin{tabular}{|c|c|c|c|c|c|c|c|c|c|}
\hline
\textbf{\% of nodes observed} & \textbf{90\%} & \textbf{80\%} & \textbf{70\%} & \textbf{60\%} & \textbf{50\%} & \textbf{40\%} & \textbf{30\%} & \textbf{20\%} & \textbf{10\%} \\
\hline

Mean & 0.77 & 0.89 & 0.91 & 0.96 & 0.97 & 0.97 & 0.99 & 0.98 & 1.11\\
\hline
Gaussian & 0.74 & 0.85 & 0.87 & 0.92 & 0.93 & 0.93& 0.95 & 0.94 & 1.06\\
\hline
kNN & 0.60 & 0.67 & 0.71 & 0.78 & 0.83 & 0.87 & 0.96 & 1.06 & 1.28\\
\hline
FUNS-N (no labels) & 0.50 & 0.55 & 0.57 & 0.61 & 0.63 & 0.64 & 0.69 & 0.73 & 0.91\\
\hline
FUNS-N & \textbf{0.45} & \textbf{0.48} & \textbf{0.51} & \textbf{0.53} & \textbf{0.55} & \textbf{0.57} & \textbf{0.59} & \textbf{0.61} & \textbf{0.73}\\

\hline
\end{tabular}
\label{tab:sumo}
\end{center}
\end{table*}

In our first experiment, we use our model to predict the live state of all unobserved locations. For that, we evaluate our model with the future time step $t=0$. This task is very similar to the task of imputation, with the special case that observations of a location are not just missing temporarily but missing permanently and only past information can be used. We describe considered baselines, that take spatial dependencies into account next.

\paragraph{Mean Prediction}
Our first trivial baseline is the prediction of the same constant value at all nodes. As a constant value, we use the mean of all training observations, independently of the analyzed time step or node.

\paragraph{k-Nearest Neighbors}
Similarly to~\cite{1041310}, we compare GNNs to a classical k-Nearest Neighbor approach.
Given our expectation of strong spatial correlation, we evaluate making a prediction based on available neighboring observations. We set $k$ adaptively to the number of available observations adjacent to each unobserved location. We calculate the mean value of these neighbors. In the case when no direct neighbors are observed, the next closest observed neighbors are used.

\paragraph{Gaussian Process Regression}
For spatial imputation of unobserved nodes, Gaussian Process Regression has been applied in previous work~\cite{liebig2012pedestrian}. We use the RBF-Kernel with $\sigma=3$ and measure pairwise distances between spatial locations of all roads. For each location, we predict the mean value of the Gaussian Process Regression. Note that this approach does not consider any temporal relations.

\paragraph{FUNS-N}
We evaluate our framework and model as described in Sec.~\ref{sec:forecasting} with $t=0$. We also evaluate the effect of the static node information by using two versions - one is using labels and one is not.
We use the same hyperparameters for FUNS-N across all experiments, only tuning the hidden dimension. We set the hidden dimension to $h=8$ and use a dropout probability of $p=0.25$ on $\mathbf{H}^{t}$ and $\mathbf{P}^{t}$ at each step during training.

We conduct experiments with the share of observed nodes used for optimization ranging from $90\%$ to $10\%$. Validation and Test nodes are set to the remaining nodes, split equally. 
The observed nodes are further split for optimization, with $50\%$ of observed nodes serving as inputs and the other $50\%$ serving as targets for the optimization. With $t=0$, we want to evaluate how effective our approach is at a task similar to matrix completion, so for this initial experiment, we do not perform a temporal data split.
Each experiment is repeated five times with different data splits and parameter initializations but kept consistent between different models.

The average mean squared error (MSE) is presented in Fig.~\ref{tab:sumo}. 
Our approach outperforms all baselines by margins to the second-best non-FUNS-N approach between $33\%$ and $61\%$. Even with only $30\%$ of the nodes observed, our FUNS-N performs better than all baselines utilizing $90\%$, meaning that reducing the number of sensors does not lead to huge drops in prediction error. 
Even when not taking static node labels into account, we find improvements of $16\%$ to $38\%$ to the best baseline. Interestingly, the gaussian process regression performed best when the predicted values are close to the mean prediction. KNN works well initially, when more than $50\%$ of the nodes are observed, though its performance declines fast when observations become sparse.

\subsubsection{Forecasting}
\begin{figure}
\centering
  \centering
  \includegraphics[width=0.5\linewidth]{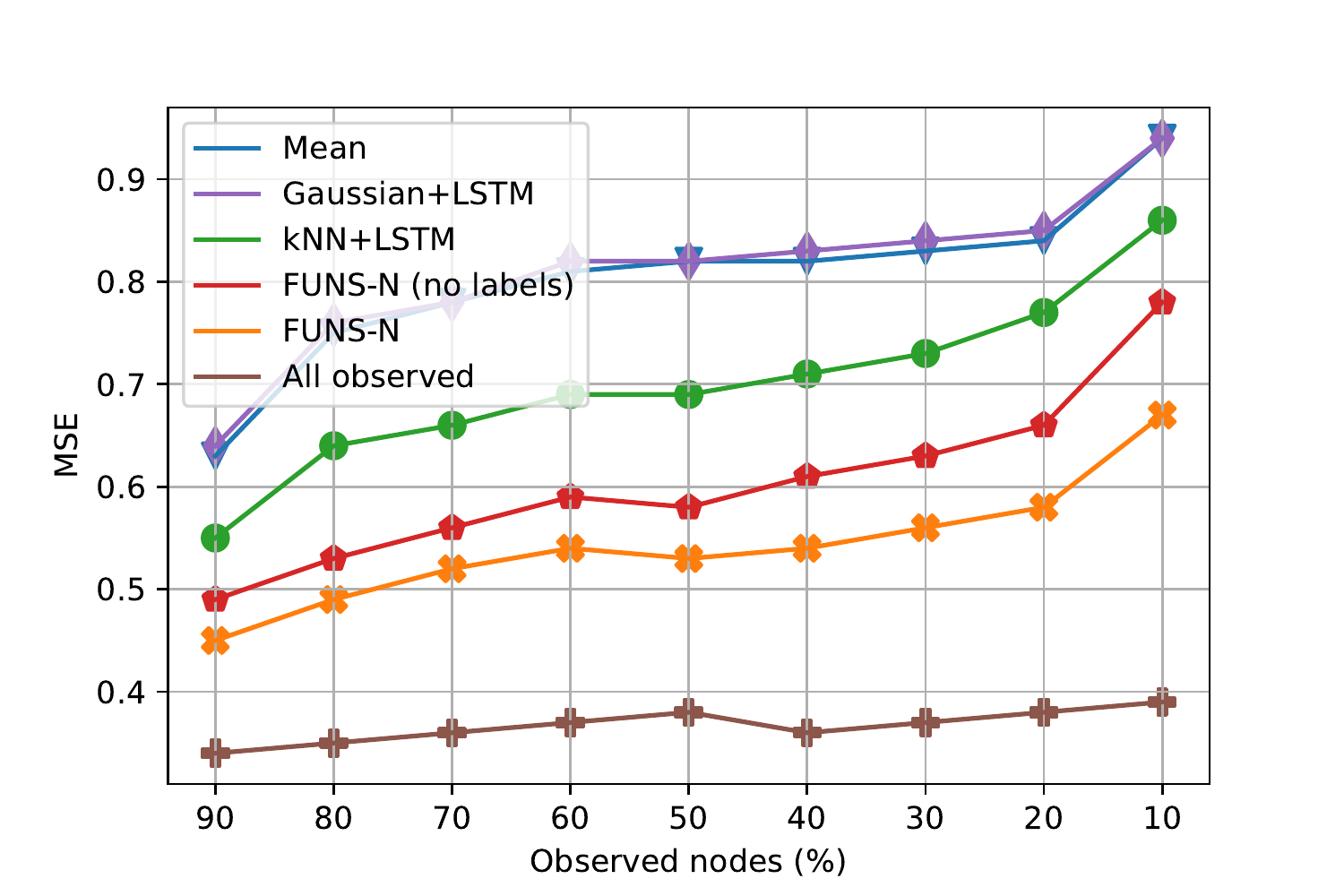}
\caption{Comparison of MSE at $t=12$ for the LuST Scenario with the share of observed nodes ranging between $90\%$ and $10\%$.}
\label{fig:results_future}
\end{figure}
We now evaluate the forecast of future states at one hour into the future, equivalently to $t=12$ for our data.
Since we are the first to consider this task, there are not baselines available and as mentioned earlier, none of the existing approaches is directly applicable here. Therefore, we combine the methods for spatial imputation with a temporal model, namely a Long short-term memory (LSTM)~\cite{10.1162/neco.1997.9.8.1735}. 
We start by using kNN and the gaussian process regression to impute the past and present state and train an LSTM for each of these methods to forecast the future state at each node.
The optimization is identical to the optimization of our FUNS-N. As a lower bound on the error, we additionally compare FUNS-N with the task of spatio-temporal forecasting for which all observations are fully available at all locations. Here, all nodes are used as inputs and as optimization targets for our FUNS-N, including validation and test nodes.

Results for the share of observed nodes ranging between $90\%$ and $10\%$ are shown in Fig.~\ref{fig:results_future}. 
The results show very similar trends to the live prediction results from Fig.~\ref{tab:sumo}. Our FUNS-N exceeds all non-FUNS-N baselines by at least $22\%$ and still performs better than all others even with only $40\%$ of nodes observed. The performance of the kNN decreases much slower for this task, also showing the effectiveness of FUNS to less powerful models. Not using static node labels for FUNS-N is again roughly $30\%$ worse in MSE for all experiments. This further emphasizes the importance of acquiring locational properties, instead of deploying and maintaining sensors.
The performance degradation with respect to the lower bound on the error with all nodes observed is reduced by up to $50\%$ compared to the baselines. Still, the error is considerably lower, when having all locations observed.
We did not find any spatial correlation between high errors.

\subsection{METR-LA}

\begin{table*}[htbp]
\caption{MSE for the forecasts of the state of unobserved nodes with $t=12$ for METR-LA}
\begin{center}
\begin{tabular}{|c|c|c|c|c|c|c|c|c|c|}
\hline
\textbf{\% of nodes observed} & \textbf{90\%} & \textbf{80\%} & \textbf{70\%} & \textbf{60\%} & \textbf{50\%} & \textbf{40\%} & \textbf{30\%} & \textbf{20\%} & \textbf{10\%} \\
\hline
Mean & 1.29 & 1.29 & 1.29 & 1.29 & 1.29 & 1.29 & 1.29 & 1.29 & 1.30\\
\hline
kNN+LSTM & 0.79 &	0.84 & 0.87 & 0.87 & 0.90 & 0.91 & 0.94 & 0.94 & 1.00 \\
\hline
FUNS-N & \textbf{0.68} & \textbf{0.74} & \textbf{0.77} & \textbf{0.77} & \textbf{0.79} & \textbf{0.80} & \textbf{0.81} & \textbf{0.85} & \textbf{0.94}\\
\hline
\hline
All observed & 0.58 & 0.58 &0.58 & 0.58 &0.58 & 0.58 &0.58 &0.58 &0.59\\
\hline
\end{tabular}
\label{tab:metrla}
\end{center}
\end{table*}

As a real-world dataset, we use METR-LA~\cite{li2018dcrnn_traffic} to forecast the traffic speed one hour into the future on highways in Los Angeles County. Since real-world datasets typically are less detailed than our simulated data, we want to examine if the state of unobserved node states can also be forecasted without a detailed graph representation and additional node labels.
METR-LA has data from $207$ loop detectors that are placed only on highways and were recorded between March and June in 2012. Only speed is available as a feature and is aggregated over $5$ minute intervals. This dataset has been used extensively as a benchmark dataset for the task of traffic forecasting before~\cite{https://doi.org/10.1111/tgis.12644,li2018dcrnn_traffic,zhang2020spatio,wu2019graph}. We use PyTorch Geometric Temporal~\cite{rozemberczki2021pytorch} as source for METR-LA.
 
Edges are constructed using the pairwise euclidean distance between locations of the loop detectors up to some threshold $\delta$. Therefore, the graph structure might not actually represent the road structure, because nodes might not actually be connected by a route and each node is connected to many others. 
There are also no static node labels available, so we can only evaluate one instance of FUNS-N. While these conditions do not use the full potential of our framework, this gives us a chance to evaluate our approach here. Experimental settings are exactly the same as for LuST, only changing the hidden dimension of FUNS-N to $h=32$.
We do not evaluate the gaussian process regression, because it has performed poorly in the previous experiments.

Results are displayed in Table~\ref{tab:metrla}. FUNS-N consistently outperforms all considered baselines by up to $16\%$. Interestingly, the difference in MSE in having only $30\%$ of the nodes observed to having $70\%$ observed is only $5\%$. 
The difference to the fully observed case is reduced notably across all experimental settings. The margin of improvement is a bit lower than for LuST, because of the aforementioned conditions in METR-LA. This dataset also only consists of highway data, making kNN less problematic as a choice.

\section{Conclusion}
\label{sec:conclusion}
Our work introduced the task of forecasting the state at locations with no available sensors that has not been considered before, because most previous methods were not applicable for this task.
We proposed FUNS, a framework than allows optimization and generalization of models based solely on the set of observed nodes, the graph inductive bias, and potentially prior knowledge about locations. Our model for this task, FUNS-N, demonstrates that GNNs are in fact the first family of models that achieve competitive results for this task. 
Our experiments on a simulated and a real-world dataset found FUNS-N to outperform all considered baselines in all settings, in some cases by more than $50\%$.
FUNS-N lowers the performance degradation, showing the potential to reduce the number of employed sensors by up to $70\%$.
As this work is only the exploration of this task, we think this may be a challenging new application domain for research on spatio-temporal models - especially for GNNs.

Our framework concurrently optimizes forecasts for observed and unobserved nodes by design, making the integration into existing forecasting models possible. Future research on the task of general forecasting could improve results for forecasting unobserved states with little extra costs.
FUNS-N largely profited from attentional sharing of information between nodes based on available observations, static node labels and other information. In future work, this could get expanded further by using different sets of parameters in the context of heterogeneous graphs~\cite{sun2013mining}.
Another promising direction is to determine the most effective locations for sensor placement. We found large differences depending on which sets were randomly sampled for optimization.

\bibliographystyle{unsrtnat}
\bibliography{bibtex}  






\end{document}